%% file: root.tex
\documentclass[letterpaper, 10 pt, conference]{ieeeconf}  

\IEEEoverridecommandlockouts                              

\usepackage{graphics} 
\usepackage{epsfig} 
\usepackage{mathptmx} 
\usepackage{times} 
\usepackage{amsmath} 
\usepackage{amssymb}  
\usepackage{times}
\usepackage{graphicx}
\makeatletter
\let\NAT@parse\undefined
\makeatother
\usepackage{multicol}
\usepackage{multirow}
\usepackage{booktabs}
\usepackage{threeparttable}
\usepackage{tabularx}
\usepackage{xcolor}
\usepackage{pifont}
\usepackage{wrapfig}
\usepackage{caption}
\usepackage{booktabs} 
\usepackage{subfigure}
\usepackage{placeins}
\usepackage{hyperref}

\title{\LARGE \bf
SP-HLM: Sparse-Pose-Driven Motion Generation and Whole-Body Control for Humanoid Loco-Manipulation
}

\author{
    \authorblockN{Zhaohui Yang$^{1}$\authorrefmark{1}, Quanbao Lin$^{1}$\authorrefmark{1}, Chengqi Shi$^{1}$\authorrefmark{1}\authorrefmark{2}, Zhenglang Weng$^{1}$\authorrefmark{2}, Jicheng Lin$^{1}$\authorrefmark{2}, Boyu Chen$^{3}$\\
    Tengjie Zhu$^{1}$, Guanzhu Ren$^{1}$, Jiahao Xu$^{2}$, Yang Feng$^{2}$, Xiaokang Yang$^{1}$, Yao Mu$^{1}$\authorrefmark{3}}
    \authorblockA{$^{1}$Shanghai Jiao Tong University \quad $^{2}$Moore Threads \\
    $^{3}$The University of Hong Kong}
    \authorblockA{\authorrefmark{1}Equal contribution. \authorrefmark{2}Interns at Shanghai Jiao Tong University. \\
    \authorrefmark{3}Corresponding author: \texttt{muyao@sjtu.edu.cn}}
}

\IEEEaftertitletext{%
    \begin{minipage}{\textwidth}
        \centering
        \includegraphics[width=\linewidth]{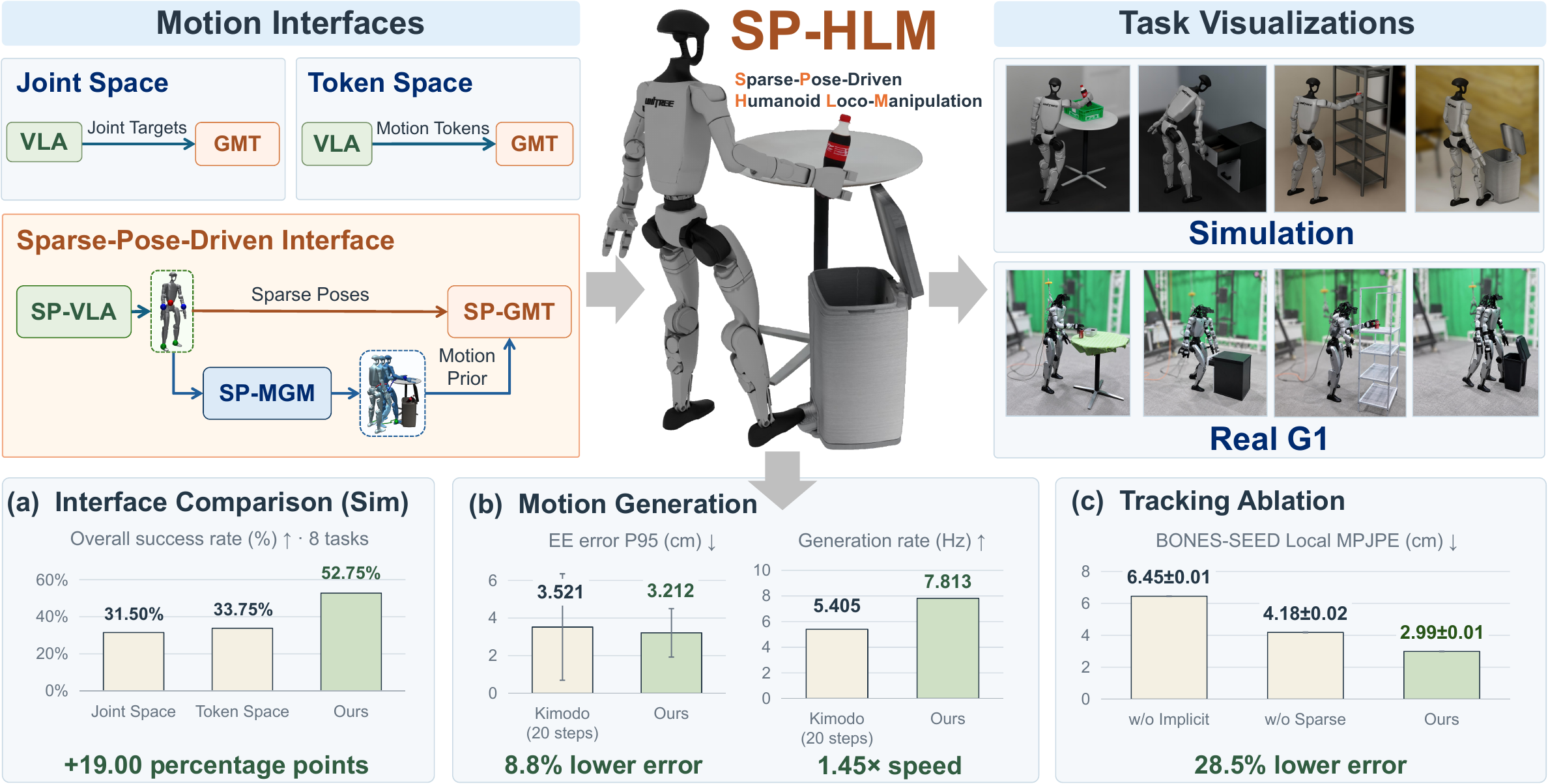}
        \captionof{figure}{\textbf{SP-HLM: Sparse-pose-driven humanoid loco-manipulation.} \textit{Top:} The interface combines explicit five-point targets with an implicit motion prior for diverse simulated and real-world tasks. \textit{Bottom:} Results show higher task success, faster motion generation, improved end-effector adherence, and lower tracking error.}
        \label{fig:sp-hlm-teaser}
    \end{minipage}%
    \vspace{1em}
}

\begin{document}

\maketitle
\thispagestyle{empty}
\pagestyle{empty}

\begin{abstract}

Humanoid loco-manipulation requires an effective interface between high-level task policies and whole-body control. Existing interfaces typically rely on a single action representation for both task intent and whole-body coordination. This may burden the high-level policy or leave one aspect underspecified. We present SP-HLM, a framework built on a sparse-pose-driven interface. The interface combines five task-critical body poses with a learned latent motion prior to represent spatial intent and whole-body coordination in complementary forms. SP-VLA, a vision-language-action model, first predicts task-relevant sparse poses and finger joint positions from multimodal observations and language instructions using asynchronous action-chunk inference. Given these sparse targets, SP-MGM, a motion generation model trained through two-stage distillation, reconstructs coordinated full-body motion in real time. A pretrained motion encoder maps this motion to a compact latent representation. SP-GMT, a general motion tracker, then integrates the explicit sparse targets with this implicit motion prior for whole-body control. Experiments in simulation and on a real humanoid robot demonstrate improved motion tracking, whole-body coordination, and loco-manipulation success over existing approaches. Additional results and videos are available on our \href{https://sp-hlm.github.io/}{\textcolor{blue}{project page}}.


\end{abstract}

\input{content/introduction}
\input{content/relatedwork}
\input{content/method}
\input{content/experiment}

\input{content/conclusion}






\bibliographystyle{IEEEtran}
\bibliography{references}
\section{Appendix}
\input{content/appendix.tex}
\end{document}

%% file: content/introduction.tex
\section{INTRODUCTION}

The motion interface between high-level task policies and whole-body control
is an important design choice in humanoid loco-manipulation
~\cite{jiang2025wholebodyvla,yin2025visualmimic, chen2026unit}. Existing interfaces emphasize different aspects of humanoid motion. Joint-space actions~\cite{hu2026openhlm,wei2026psi0, xiao2026rove} directly specify joint configurations, but can require the high-level policy to model whole-body coordination alongside task intent. Task-space actions provide a compact, task-grounded representation, yet sparse targets alone do not fully specify coordinated whole-body motion. Token-space actions~\cite{luo2026sonic} can encode reusable coordination patterns, while task-critical spatial targets may remain implicit. These trade-offs motivate an interface that combines representations of spatial intent and whole-body coordination, rather than encoding both in a single form.

We argue that task-critical spatial intent and whole-body coordination should instead be represented in complementary forms. Spatial intent remains explicit, specifying \emph{where} task-relevant body parts should move, whereas whole-body coordination is encoded implicitly as a learned motion prior describing \emph{how} the body should move. Following this principle, we expose only five task-critical body poses---the pelvis, two wrists, and two ankles---as explicit spatial constraints and complement them with a latent whole-body motion representation. This interface preserves direct task-space control while avoiding the need for the high-level policy to explicitly specify full-body coordination.

Following this formulation, we present SP-HLM, a humanoid loco-manipulation framework built on a sparse-pose-driven interface. At its core, SP-GMT performs whole-body control by jointly conditioning on sparse task-space targets and a learned motion latent. We pretrain a motion variational autoencoder (VAE) to encode full-body motion into a compact latent representation. This latent provides coordination information not fully specified by the sparse targets while enabling efficient policy training.

To obtain this implicit motion prior from sparse task-space targets at deployment, we introduce SP-MGM, which reconstructs coordinated full-body motion from five-point pose targets in real time. Starting from a pretrained motion generation model, SP-MGM reduces network depth and denoising steps and employs two-stage distillation to improve sparse-pose adherence and temporal consistency.

The proposed interface also reduces the burden of high-level action prediction. SP-VLA predicts only task-relevant sparse body poses and finger joint positions from multimodal observations and language instructions, without directly predicting full-body configurations. We further adopt asynchronous action-chunk inference for responsive closed-loop execution. Together, SP-VLA specifies task-space intent, SP-MGM recovers coordinated whole-body motion, and SP-GMT integrates the explicit sparse targets and implicit motion prior for whole-body control.

We separately evaluate SP-GMT, SP-MGM, and the complete SP-HLM system. SP-GMT achieves competitive motion-tracking performance against representative state-of-the-art baselines. SP-MGM improves sparse-pose adherence over the motion-generation baseline while achieving a 1.45$\times$ higher generation rate. For loco-manipulation, SP-HLM achieves a 52.75\% overall success rate across eight simulated tasks, outperforming the token-space and joint-space baselines. We further validate SP-HLM on four real-world tasks, demonstrating its feasibility for humanoid loco-manipulation. Our main contributions are:
\begin{itemize}
    \item We propose a sparse-pose-driven motion interface for humanoid loco-manipulation, combining five task-critical body poses with a learned motion prior to represent spatial intent and whole-body coordination in complementary forms.

    \item We develop \textbf{SP-GMT}, a whole-body tracker conditioned jointly on sparse targets and a pretrained motion representation, and \textbf{SP-MGM}, a real-time sparse-conditioned motion generator trained through two-stage distillation.

    \item We present \textbf{SP-HLM}, a humanoid loco-manipulation framework that integrates the above two components with \textbf{SP-VLA}, which uses asynchronous action-chunk inference to predict task-relevant sparse poses and finger joint-position targets from multimodal inputs. The resulting system achieves higher overall success than the evaluated baselines across eight simulated tasks and is further validated on four real-world tasks.

\end{itemize}

%% file: content/relatedwork.tex
\section{RELATED WORK}

\subsection{General Humanoid Motion Generation and Control}

Recent work increasingly uses motion generation as an intermediate layer between high-level semantic commands and low-level humanoid control, typically producing dense whole-body references for downstream tracking~\cite{xie2026textop,jiang2025uniact,huang2026omg}. Some methods formulate motion synthesis as an in-betweening problem~\cite{wang2026motionbricks}, whereas others employ generative models as online motion planners~\cite{zhang2026wholebody,chen2026reactivebfm}. Although these approaches demonstrate the effectiveness of generative motion priors for hierarchical control, they largely rely on dense motion references rather than explicit task-space interfaces tailored to loco-manipulation.

In parallel, general motion tracking has emerged as a scalable approach to acquiring versatile whole-body control from large and diverse motion corpora~\cite{luo2026sonic,chen2025gmt,li2026omnitrack,zhang2025track}. Recent studies improve generalization by scaling motion data and model capacity~\cite{qi2026generative,qi2026humanoidgpt,chen2026holomotion,zeng2026scaling}, while others leverage heterogeneous motion sources or expert specialization~\cite{sleiman2026zest,pan2025agility,wang2026experts}. Related efforts learn reusable behavioral representations that can be transferred to downstream control tasks~\cite{li2026bfm,liao2025beyondmimic}. Together, these advances provide increasingly general and robust motion generation and control capabilities, while the interface for conveying task-relevant whole-body intent from high-level loco-manipulation policies remains comparatively less explored.

\subsection{Motion Interfaces for Humanoid Loco-Manipulation}
\label{sec:related_hlm}

Recent humanoid loco-manipulation systems increasingly pursue unified whole-body behaviors through vision-language-action models (VLAs)~\cite{jiang2025wholebodyvla,wei2026psi0,hu2026openhlm,gu2026haf,liu2025trajbooster}, world-action models~\cite{zheng2026motionwam,li2026omega}, and perceptive reinforcement policies~\cite{yin2025visualmimic,he2025viral}. Beyond the choice of policy architecture, an important design consideration is the \emph{motion interface} through which task-level intent is conveyed to whole-body execution. Existing approaches span decoupled upper- and lower-body control~\cite{wei2026psi0,jiang2025wholebodyvla} and explicit whole-body kinematic representations~\cite{wang2026vlk,hu2026openhlm}. While these interfaces provide direct control over robot motion, they can require the high-level policy to reason over relatively detailed body configurations. Other methods adopt more compact learned motion representations, such as motion tokens~\cite{zheng2026motionwam,xue2025leverb}, which capture reusable coordination patterns but make task-relevant spatial constraints less explicit. 

Recent work also explores compact task-space interfaces based on sparse body targets or low-rate task-space trajectories, leaving the remaining whole-body coordination to downstream controllers~\cite{yang2026handoff,ge2026cwi,liu2026learning}. In contrast, SP-HLM combines explicit task-space targets with a learned whole-body motion prior, providing a compact interface between task-level intent and coordinated whole-body execution.

%% file: content/method.tex
\section{METHOD}
\label{sec:method}

\begin{figure*}[!t]
    \vspace*{2mm}
    \centering
    \includegraphics[width=\textwidth]{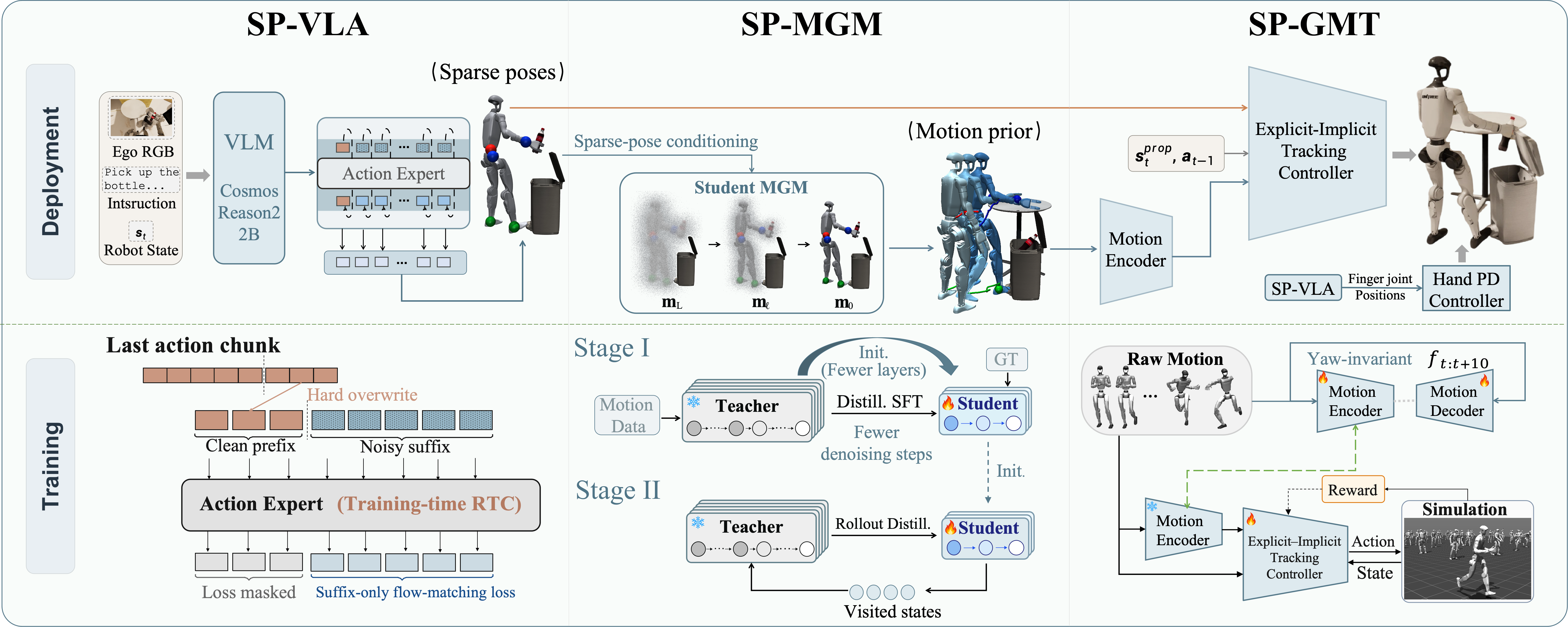}
    \caption{Overview of the proposed SP-HLM framework based on the sparse-pose-driven interface. \textbf{Top:} the deployment pipeline integrating sparse-pose prediction, full-body motion generation, and explicit--implicit whole-body control. \textbf{Bottom:} the training procedures for SP-VLA, SP-MGM, and SP-GMT.}
    \label{fig:sp-hlm-overview}
\end{figure*}

\subsection{SP-HLM Overview}

As illustrated in Fig.~\ref{fig:sp-hlm-overview},
SP-HLM connects high-level loco-manipulation policies
to low-level whole-body control through a sparse-pose-driven interface.
The sparse poses specify task-relevant spatial constraints
on five body parts in the world frame: the pelvis, both
wrists, and both ankles.
Each pose comprises a 3-D position and a 6-D rotation.\footnote{The 6-D representation is obtained
by flattening the first two columns of the rotation matrix
in row-major order:
$[R_{11}, R_{12}, R_{21}, R_{22}, R_{31}, R_{32}]$.}
In addition,
a full-body motion prior generated from these poses provides
whole-body coordination consistent with the specified task
geometry.
An explicit--implicit tracking policy then integrates
the sparse-pose constraints with the latent motion prior
to enable humanoid loco-manipulation.

Specifically, at each planning step, SP-VLA predicts a
sequence of five-point target poses together with the finger joints of both hands.
SP-MGM then generates a full-body motion trajectory
conditioned on these targets, and a pretrained motion
encoder maps the trajectory to a latent command.
SP-GMT combines this latent command with the sparse targets
expressed in the current pelvis frame to produce joint position commands. The finger targets are passed directly to the hand PD
controllers.
Formally,
\begin{equation}
\begin{aligned}
\boldsymbol{m}_{t:t+H}
&= G_{\psi}(\mathbf{k}^{W}_{t:t+H}),
&\quad
\mathbf{z}_t
&= E_{\phi}(\boldsymbol{m}_{t:t+H}),\\
\mathbf{k}^{b}_t
&= \mathcal{T}_{P_t}(\mathbf{k}^{W}_t),
&\quad
\mathbf{a}_t
&= \pi_{\theta}\bigl(
\mathbf{z}_t,
\mathbf{k}^{b}_t,
\mathbf{s}^{\mathrm{prop}}_t,
\mathbf{a}_{t-1}
\bigr),
\end{aligned}
\label{eq:sohlm-overview}
\end{equation}
where $\mathbf{k}^{W}_{t:t+H}$ denotes the five-point target
sequence from SP-VLA; $G_{\psi}$, $E_{\phi}$, and
$\pi_{\theta}$ denote SP-MGM, the motion VAE encoder, and
the SP-GMT policy, respectively.
The notation is explained in detail in the following sections.

\subsection{SP-GMT: Explicit--Implicit General Motion Tracking}
\label{sec:gmt}

SP-GMT implements our sparse-pose-driven interface at the low-level control stage through explicit-implicit integration.
Sparse body poses explicitly specify task-critical spatial
constraints, while the motion latent encodes whole-body
coordination. The policy combines both signals to enable precise
and dynamically feasible whole-body control.

\subsubsection{Yaw-Invariant Latent Motion Prior}
\label{sec:motion-representation}

We pretrain a motion variational autoencoder (VAE) on approximately 190 hours
of motion data to learn a representation of coordinated whole-body
motion. Let $J$ denote the number of actuated joints. Each motion frame is
represented as
\begin{equation}
\mathbf{f}_t =
\left[
\mathbf{q}_t,\,
\dot{\mathbf{q}}_t,\,
\mathbf{r}^{b}_{p,t},\,
\mathbf{v}^{b}_{p,t},\,
\boldsymbol{\omega}^{b}_{p,t}
\right]
\in \mathbb{R}^{2J+12},
\label{eq:motion-state}
\end{equation}
where $\mathbf{q}_t$ and $\dot{\mathbf{q}}_t$ are joint positions and
velocities, $\mathbf{r}^{b}_{p,t}$ is the 6-D pelvis orientation, and
$\mathbf{v}^{b}_{p,t}$ and $\boldsymbol{\omega}^{b}_{p,t}$ are pelvis linear
and angular velocities. Notably, all pelvis quantities are expressed in a frame aligned with the current
pelvis yaw, while global horizontal translation and heading are removed.
The resulting representation is invariant to global placement while
preserving articulated pose, pelvis tilt, and local root dynamics, facilitating compact motion encoding.

At each time step, we construct a short motion window
\begin{equation}
\mathbf{F}_t =
\left[
\mathbf{f}_{t+1},\ldots,\mathbf{f}_{t+H}
\right],
\qquad H=10,
\label{eq:motion-window}
\end{equation}
and encode it using a skip-connected transformer. The VAE is trained by minimizing
\begin{equation}
L_{\mathrm{VAE}}
=
L_{\mathrm{rec}}
+
\beta
D_{\mathrm{KL}}
\left(
E_{\phi}(\mathbf{z}_t|\mathbf{F}_t)
\parallel
\mathcal{N}(\mathbf{0},\mathbf{I})
\right),
\label{eq:vae-loss}
\end{equation}
where $L_{\mathrm{rec}}$ equally weights reconstruction errors for
joint states, pelvis orientation, and pelvis velocities, and $\beta$ controls the KL regularization strength. 

After pretraining, the
encoder is frozen and its posterior mean is used as the deterministic motion
latent.
This latent provides a compact whole-body motion
representation for the downstream tracking policy,
which also helps improve training efficiency. 
VAE training is conducted on Moore Threads S5000 GPUs.

\subsubsection{Explicit--Implicit Integration}
\label{sec:explicit-implicit-interface}

SP-GMT integrates an implicit motion latent with explicit
sparse-pose targets:
\begin{equation}
\mathbf{z}_t = E_{\phi}(\boldsymbol{m}_{t:t+H}),
\qquad
\mathbf{k}^{b}_t =
\mathcal{T}_{P_t}(\mathbf{k}^{W}_t).
\label{eq:explicit-implicit-interface}
\end{equation}

For the implicit branch, $E_{\phi}$ maps the full-body
motion $\boldsymbol{m}_{t:t+H}$ to $\mathbf{z}_t$ through
yaw-aligned preprocessing and motion encoding.
For the explicit branch, $\mathcal{T}_{P_t}$ transforms
the world-frame five-point poses $\mathbf{k}^{W}_t$
into the current robot pelvis frame.
Each target is represented as
$\mathbf{k}^{b}_{t,i}
=[\mathbf{p}^{b}_{t,i},\,\mathbf{r}^{b}_{t,i}]$,
where $\mathbf{p}^{b}_{t,i}\in\mathbb{R}^{3}$ and
$\mathbf{r}^{b}_{t,i}\in\mathbb{R}^{6}$ denote its relative
position and orientation. 

In addition to these two commands, the policy receives the
current proprioceptive state $\mathbf{s}^{\mathrm{prop}}_t$
and previous action $\mathbf{a}_{t-1}$.
The resulting actor observation is
$\mathbf{o}^{\pi}_t=
[\mathbf{z}_t,\mathbf{k}^{b}_t,
\mathbf{s}^{\mathrm{prop}}_t,\mathbf{a}_{t-1}]$,
from which the policy predicts body joint-position targets.
Here, $\mathbf{s}^{\mathrm{prop}}_t$ comprises the body joint
positions and velocities, world-frame pelvis pose, base
linear and angular velocities, and projected gravity. 
Regarding command sources, 
SP-GMT training uses five-point poses and full-body motion
from reference data, whereas SP-HLM deployment obtains them
online from SP-VLA and SP-MGM, respectively.

\subsubsection{Tracking Policy Optimization}
\label{sec:tracking-setup}

Following standard motion-tracking practice~\cite{liao2025beyondmimic}, we formulate SP-GMT training
as a goal-conditioned reinforcement learning problem and
optimize the policy with PPO.
We adopt an asymmetric actor--critic architecture, where
the critic receives privileged full-body state information during training.

The tracking reward is given by
\begin{equation}
r_t =
w_{\mathrm{sp}} r_{\mathrm{sp}}
+
w_{\mathrm{pose}} r_{\mathrm{pose}}
+
w_{\mathrm{vel}} r_{\mathrm{vel}}
-
w_{\mathrm{reg}} r_{\mathrm{reg}},
\label{eq:tracking-reward}
\end{equation}
where $r_{\mathrm{pose}}$ and $r_{\mathrm{vel}}$ encourage full-body pose and
velocity tracking, respectively, while $r_{\mathrm{reg}}$ penalizes undesirable control behavior. The additional sparse-pose term $r_{\mathrm{sp}}$ rewards the
global position and orientation tracking of the five-point poses,
thereby enforcing task-relevant spatial constraints. 

During training, we combine failure-aware and uniform
motion sampling to emphasize challenging segments while
maintaining motion coverage. We also randomize motor dynamics,
contact properties, external perturbations, action delays,
and command-filtering parameters to improve deployment robustness.

\subsection{SP-MGM: Real-Time Motion Generation from Sparse Pose Constraints}
\label{sec:motion-distillation}

SP-MGM generates the full-body motion prior for the
sparse-pose-driven interface, conditioned on five-point targets. To enable real-time inference, we compress a pretrained diffusion-based motion generator~\cite{rempe2026kimodo}
by reducing its network depth and denoising steps. A two-stage distillation strategy is then designed to
improve adherence to the sparse-pose constraints while
maintaining motion smoothness and full-body coordination.

\subsubsection{Offline Supervised Fine-Tuning (Stage I)}
The compressed student is jointly trained using frozen-teacher
predictions and paired ground-truth motions to acquire general
full-body motion generation capabilities.
The student adopts an $N_S$-step denoising schedule subsampled from
the teacher's $N_T$-step schedule. For a sampled student timestep $\tau$,
$\rho(\tau)$ denotes its noise-matched teacher timestep. The noisy motion $\mathbf{m}_{\tau}$ is constructed from the paired clean motion $\mathbf m_0$ as
\begin{equation}
\quad
\mathbf{m}_{\tau}
=
\alpha_{\tau}\mathbf{m}_0
+
\sigma_{\tau}\boldsymbol{\epsilon},
\boldsymbol{\epsilon}\sim\mathcal{N}(\mathbf{0},\mathbf{I}),
\label{eq:mgm-forward-noising}
\end{equation}
where $(\alpha_{\tau},\sigma_{\tau})$ specifies the noise level shared by student timestep
$\tau$ and teacher timestep $\rho(\tau)$. 

Given the same noisy motion and sparse
constraints, the student and teacher predict the clean motion:
\begin{equation}
\widehat{\mathbf{m}}_0^S
=
G_S\left(
\mathbf{m}_{\tau},
\mathbf{K}_t^W,
\tau
\right),
\quad
\widehat{\mathbf{m}}_0^T
=
G_T\left(
\mathbf{m}_{\tau},
\mathbf{K}_t^W,
\rho(\tau)
\right).
\label{eq:mgm-denoising-prediction}
\end{equation}
The student is then optimized with the joint objective:
\begin{equation}
\begin{aligned}
L_{\mathrm{sft}}
={}&
w_T L_{\mathrm{mot}}
\left(
\widehat{\mathbf{m}}_0^S,
\widehat{\mathbf{m}}_0^T
\right)
+
w_{\mathrm{GT}} L_{\mathrm{mot}}
\left(
\widehat{\mathbf{m}}_0^S,
\mathbf{m}_0
\right),
\end{aligned}
\label{eq:offline-finetuning}
\end{equation}
where $L_{\mathrm{mot}}$ denotes the full-body reconstruction loss over root and joint kinematics, foot contacts, and FK consistency; $w_T$ and $w_{\mathrm{GT}}$ dynamically weight the teacher and
ground-truth terms to balance the pretrained motion prior with motion reconstruction fidelity.

\subsubsection{Rollout-Aware Distillation (Stage II)}

Stage I trains the student only on forward-diffused states,
causing error accumulation and distribution shift during inference. To address
this, Stage II further applies frozen-teacher supervision to student-visited
states under five-point conditioning. After each student rollout, we sample intermediate states
$\mathbf{m}_{\tau}^S$ and obtain corrective clean-motion targets from the teacher:
\begin{equation}
\widehat{\mathbf{m}}_{0,\mathrm{corr}}^T
=
G_T
\left(
\mathbf{m}^{S}_{\tau},
\mathbf{K}_t^W,
\rho(\tau)
\right).
\label{eq:rollout-teacher-correction}
\end{equation}

To further enforce temporal smoothness, the final rollout-aware objective is:
\begin{equation}
\begin{aligned}
L_{\mathrm{roll}}
={}&
L_{\mathrm{mot}}
\left(
\widehat{\mathbf{m}}_0^S,
\widehat{\mathbf{m}}_{0,\mathrm{corr}}^T
\right)
+
\lambda_{\mathrm{temp}}L_{\mathrm{temp}},
\end{aligned}
\label{eq:online-finetuning-loss}
\end{equation}
where $L_{\mathrm{temp}}$ applies finite-difference regularization to joint
motion, pelvis translation, and orientation to suppress high-frequency variations. Through the two-stage distillation strategy, the resulting SP-MGM generates
high-fidelity, temporally smooth full-body motion using a compressed
network and fewer denoising steps.


\subsection{SP-VLA: Asynchronous Sparse-Pose and Hand-Command Prediction}
\label{sec:sp-vla}
\begin{figure*}[!tp]
    \vspace*{2mm}
    \centering
    \newcommand{\taskkeyframes}[3]{%
        \begin{minipage}[t]{0.492\textwidth}
            \vspace{0pt}
            \setlength{\fboxsep}{1.5pt}%
            \colorbox{black!15}{%
                \parbox{\dimexpr\linewidth-2\fboxsep\relax}{%
                    \raggedright\fontsize{7}{8}\selectfont\bfseries
                    Task #1: #2\par}}\par
            \nointerlineskip
            \includegraphics[width=\linewidth]{#3}%
        \end{minipage}%
    }
    \taskkeyframes{1}{Walk forward and pick up the cylinder.}
        {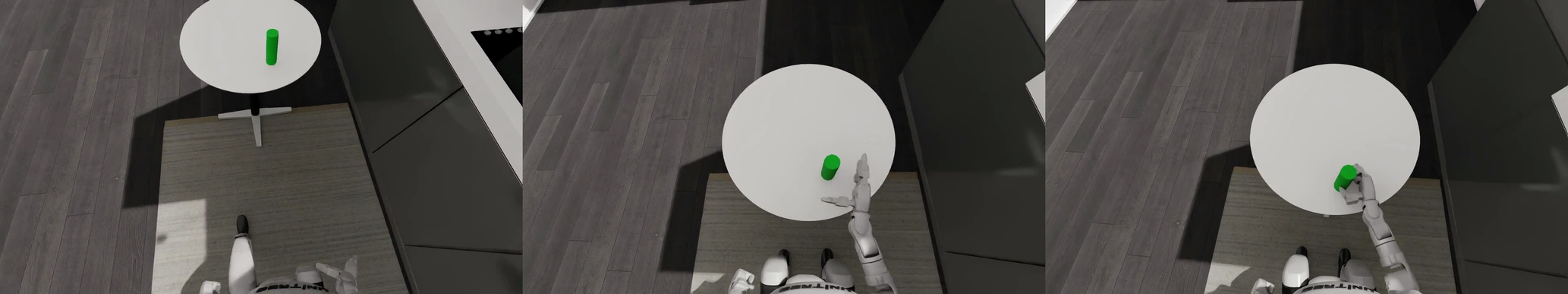}
    \hfill
    \taskkeyframes{2}{Pick up the bottle, step on the pedal to open the bin, and put it inside.}
        {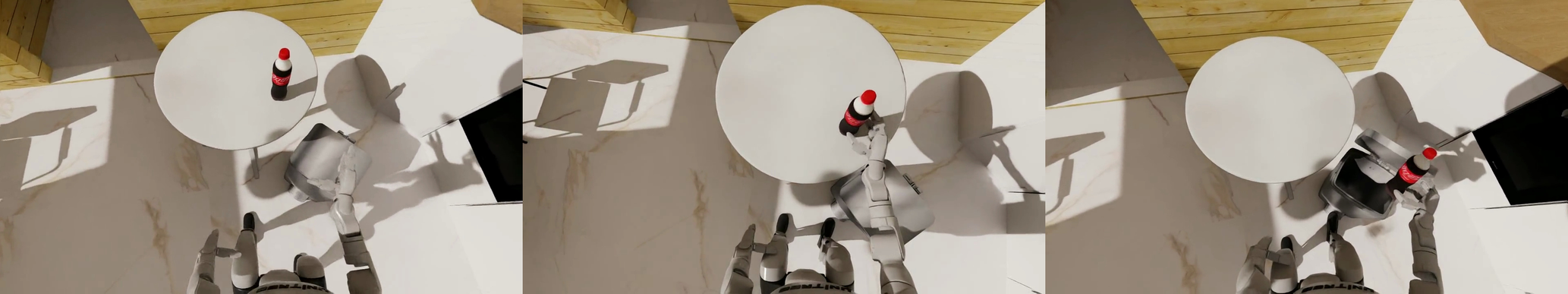}

    \par\vspace{4pt}

    \taskkeyframes{3}{Step on the pedal to open the trash bin.}
        {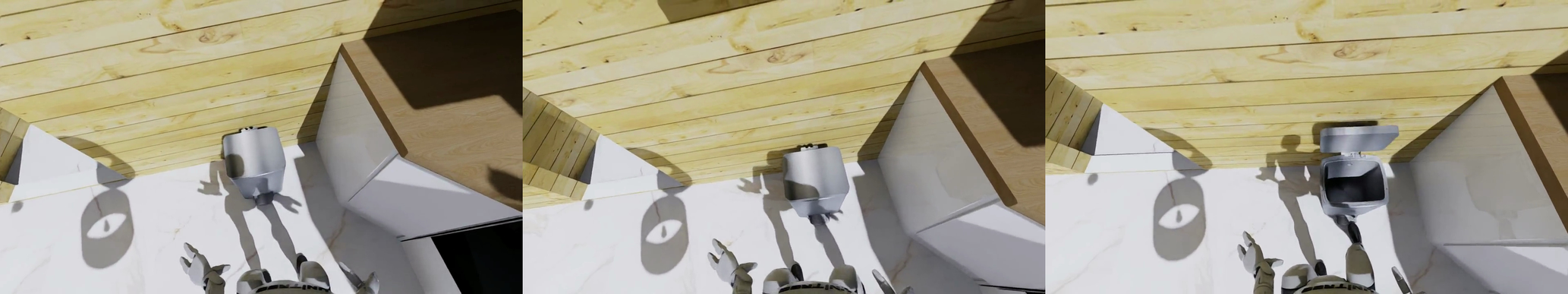}
    \hfill
    \taskkeyframes{4}{Pick up the bottle and place it in the box.}
        {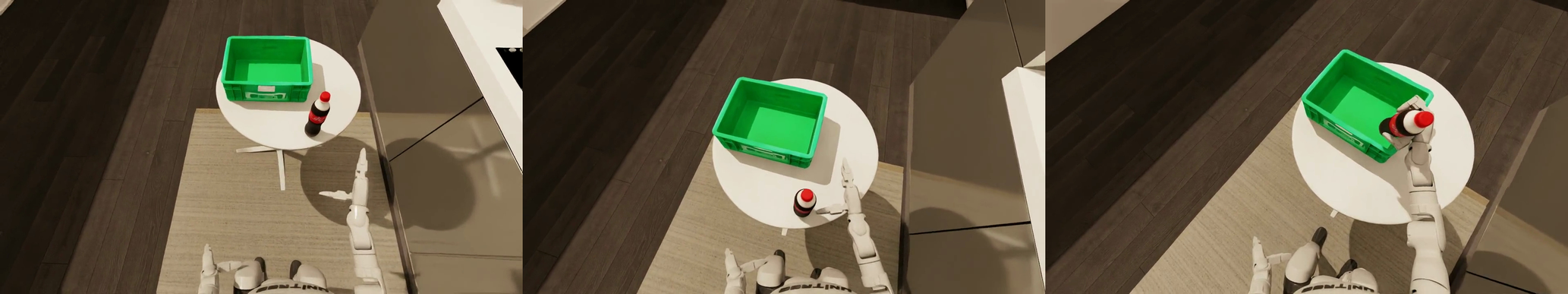}

    \par\vspace{4pt}

    \taskkeyframes{5}{Squat down and pull open the upper drawer of the cabinet.}
        {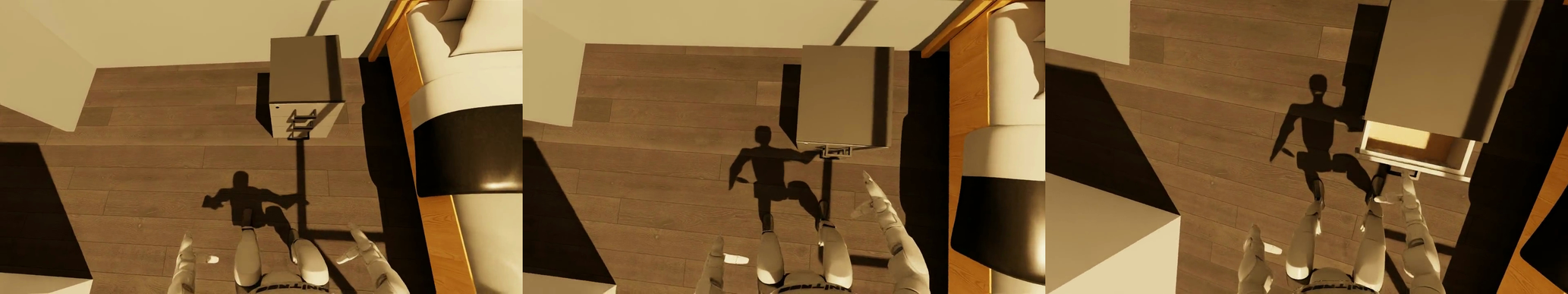}
    \hfill
    \taskkeyframes{6}{Pick up the object, squat down, and place it on the lower shelf.}
        {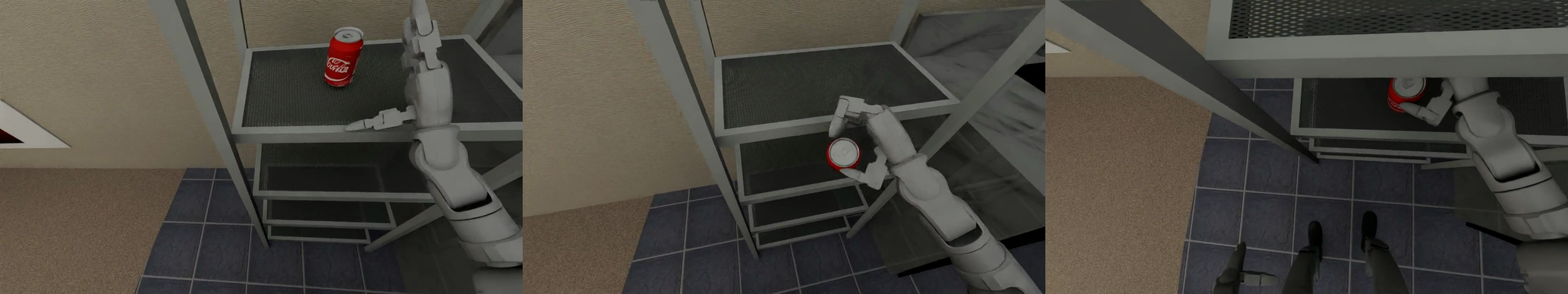}

    \par\vspace{4pt}

    \taskkeyframes{7}{Kick the football into the goal.}
        {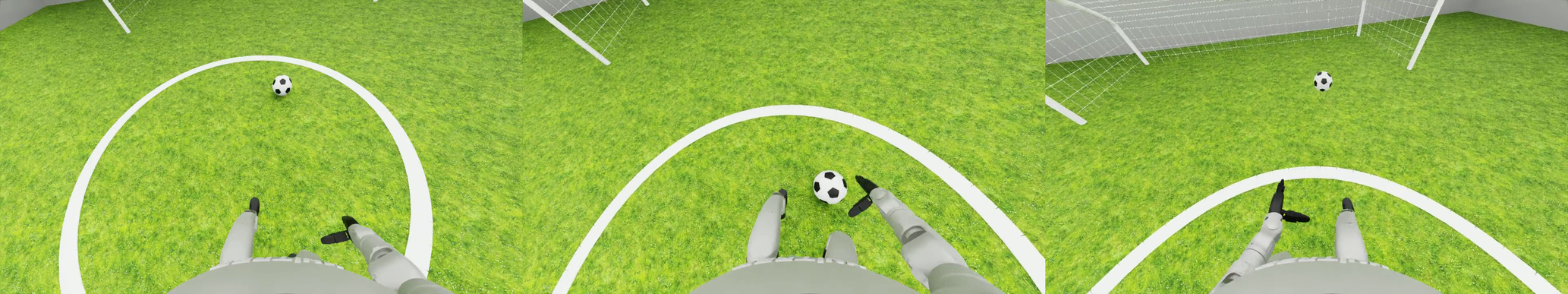}
    \hfill
    \taskkeyframes{8}{Push down the handle and push the door open.}
        {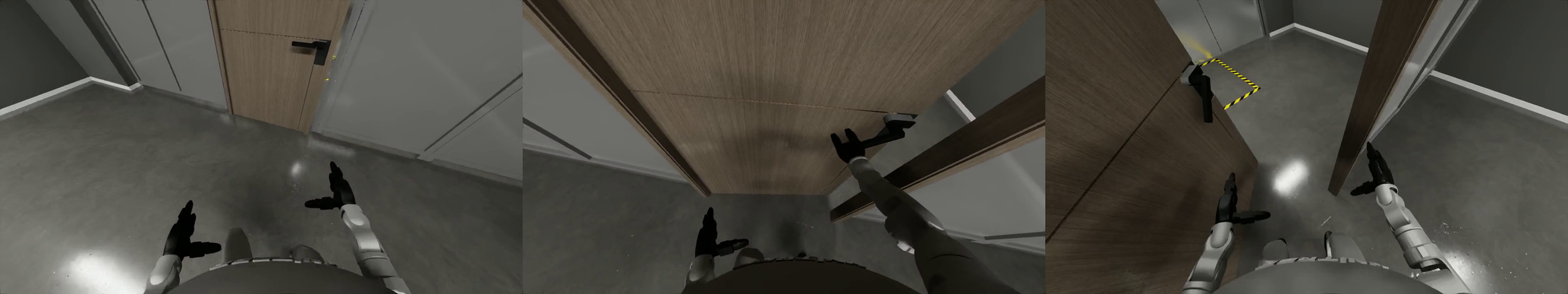}
    \caption{Qualitative rollouts of SP-HLM on eight simulated loco-manipulation tasks. Each sequence shows three temporally ordered keyframes illustrating task progression, with the corresponding language instruction shown above.}
    \label{fig:simulated-task-keyframes}
    \par\vspace{8pt}
    \captionof{table}{Success rates across eight simulated loco-manipulation tasks.}
    \label{tab:real-world-task-success-horizontal}
    \label{tab:sim_success}
    \footnotesize
    \setlength{\tabcolsep}{2pt}
    \begin{tabular*}{\textwidth}{@{\hspace{4pt}\extracolsep{\fill}}lccccccccc@{\hspace{4pt}}}
        \toprule
        Method & \shortstack{T1:\\Pick} & \shortstack{T2:\\Pick~\&~Step}
        & \shortstack{T3:\\Step} & \shortstack{T4:\\Pick~\&~Place}
        & \shortstack{T5:\\Squat~\&~Pull}
        & \shortstack{T6:\\Pick~\&~Squat}
        & \shortstack{T7:\\Kick} & \shortstack{T8:\\Push} & Overall \\
        \midrule
        \rule{0pt}{2.5ex}GR00T~\&~Token Space
        & 22/50 & 7/50 & 24/50 & 20/50
        & 8/50 & 9/50 & 19/50 & 26/50
        & 33.75\% \\[0.2ex]
        \rule{0pt}{2.5ex}GR00T~\&~Joint Space
        & 11/50 & 11/50 & 20/50 & 19/50
        & 8/50 & 21/50 & 13/50 & 23/50
        & 31.50\% \\[0.2ex]
        \rule{0pt}{2.5ex}SP-HLM (Direct Latent Prediction)
        & 20/50 & 6/50 & 20/50 & 6/50
        & 16/50 & 6/50 & 6/50 & 13/50
        & 23.25\% \\[0.2ex]
        \rule{0pt}{2.5ex}SP-HLM (Ours)
        & \textbf{26/50} & \textbf{13/50} & \textbf{39/50} & \textbf{24/50}
        & \textbf{19/50} & \textbf{26/50} & \textbf{28/50} & \textbf{36/50}
        & \textbf{52.75\%} \\[0.2ex]
        \bottomrule
    \end{tabular*}
    \vspace{2pt}

    \begin{minipage}{0.98\textwidth}
        \footnotesize
        \textit{Task descriptions.}
        \textbf{T1:} walk forward and pick up the cylinder;
        \textbf{T2:} pick up the bottle, step on the pedal to open the bin, and put it inside;
        \textbf{T3:} step on the pedal to open the trash bin;
        \textbf{T4:} pick up the bottle and place it in the box;
        \textbf{T5:} squat down and pull open the upper drawer of the cabinet;
        \textbf{T6:} pick up the object, squat down, and place it on the lower shelf;
        \textbf{T7:} kick the football into the goal; and
        \textbf{T8:} push down the handle and push the door open.
    \end{minipage}
\end{figure*}
Serving as the front end of the sparse-pose-driven interface, SP-VLA is
initialized from pretrained GR00T N1.7~\cite{bjorck2025gr00t}. We fine-tune only its
flow-matching action head while keeping the
vision--language backbone frozen.
Given an egocentric RGB image $\mathbf{o}_t$, language instruction $\ell$,
and robot state $\mathbf{s}_t$, SP-VLA predicts an $H$-step sequence of
world-frame target poses for the pelvis, both wrists, and both ankles, together
with finger joint-position targets for both hands:
\begin{equation}
\left[
\mathbf{k}_{t:t+H-1}^{W},
\mathbf{g}_{t:t+H-1}
\right]
\sim
\pi_{\eta}\!\left(
\cdot \mid \mathbf{o}_t,\ell,\mathbf{s}_t
\right),
\label{eq:sp-vla-policy}
\end{equation}
where $\mathbf{s}_t$ comprises the body joint positions,
the world-frame pelvis pose, and the finger joint positions
of both hands; $\mathbf{g}_{t:t+H-1}$ denotes the finger joint-position target sequence.

During fine-tuning, we adopt training-time RTC~\cite{black2025training}
to accommodate inference latency.
For a sampled delay $d$, let
$\mathbf{u}_i=
[\mathbf{k}^{W}_{t+i},\mathbf{g}_{t+i}]$
denote the $i$-th element of the combined action chunk.
At flow time $\tau$, we retain a clean prefix and perturb
only the suffix:
\begin{equation}
\widetilde{\mathbf{u}}^{\tau}_i =
\begin{cases}
\mathbf{u}_i, & i<d,\\
(1-\tau)\boldsymbol{\epsilon}_i+\tau\mathbf{u}_i,
& i\geq d,
\end{cases}
\qquad
\boldsymbol{\epsilon}_i\sim\mathcal{N}(\mathbf{0},\mathbf{I}),
\label{eq:rtc-prefix}
\end{equation}
where $i=0,\ldots,H-1$ and $\tau\in[0,1]$.
The flow-matching loss is applied only to the suffix,
training the policy to generate subsequent sparse poses
and hand commands conditioned on the committed prefix.
This prefix conditioning promotes temporal continuity
across action chunks.

During SP-HLM inference, SP-VLA begins generating the next
action chunk from updated observations while the current
chunk is still being executed.
This asynchronous execution reduces inference-induced waiting between
chunks. With TensorRT acceleration, SP-VLA and SP-MGM achieve inference
rates of $25\,\mathrm{Hz}$ and $10\,\mathrm{Hz}$, respectively.

%% file: content/experiment.tex
\section{EXPERIMENTS}

Through our experiments, we aim to answer the following three research questions:
\begin{itemize}
\item \textbf{RQ1:} Does the proposed explicit--implicit tracking policy improve motion-tracking performance, and how does it compare with SOTA general motion trackers?

\item \textbf{RQ2:} Can the proposed two-stage distillation strategy accelerate motion generation while preserving accurate adherence to sparse end-effector constraints?

\item \textbf{RQ3:} Does our sparse-pose-driven interface provide higher task success than joint-based and token-based interfaces across diverse loco-manipulation tasks, and how well does the complete system transfer to real-world humanoid deployment?

\end{itemize}

\subsection{Motion Tracking Performance}
We evaluate tracking on two 500-motion datasets: Filtered CLOT-500~\cite{zhu2026clot} and BONES-SEED Diverse-500~\cite{rempe2026kimodo}. Each dataset is evaluated over ten seeded runs, where perturbations are applied to the reference trajectories. Both are unseen during SP-GMT training, while BONES-SEED Diverse-500 is included in SONIC's training data.
\subsubsection{Comparison with Motion-Tracking Baselines}

\begin{table*}[htbp]
    \vspace*{1.5mm}
    \centering
    \caption{Motion tracking under trajectory observation noise over ten seeded runs.
    All methods use the same evaluation configuration. Values report
    the population mean $\pm$ standard deviation.
    Bold and underlined means indicate the best and second-best results,
    respectively.}
    \label{tab:motion-tracking-trajectory-noise}

    \footnotesize
    \setlength{\tabcolsep}{3.2pt}
    \renewcommand{\arraystretch}{1.08}

    \begin{tabular*}{\textwidth}{@{\extracolsep{\fill}}lcccccc@{}}
        \toprule
        \multicolumn{1}{c}{Method}
        & Progress $\uparrow$
        & Local MPJPE $\downarrow$
        & Global MPJPE $\downarrow$
        & Joint error $\downarrow$
        & EE global error $\downarrow$
        & EE local error $\downarrow$ \\
        & (\%) & (m) & (m) & (rad) & (m) & (m) \\
        \midrule

        \multicolumn{7}{l}{\textit{Filtered CLOT-500}} \\

        SP-GMT w/ Joint-Space Reference

        & $99.35 \pm 0.14$
        & $0.0305 \pm 0.0000$
        & ${0.0812} \pm 0.0004$
        & $0.0975 \pm 0.0001$
        & ${0.0879} \pm 0.0004$
        & $0.0571 \pm 0.0001$ \\

        SP-GMT w/o Sparse Reference
        & $99.28 \pm 0.19$
        & $0.0390 \pm 0.0000$
        & $0.4436 \pm 0.0036$
        & $0.1538 \pm 0.0000$
        & $0.4506 \pm 0.0036$
        & $0.0733 \pm 0.0000$ \\

        SP-GMT w/o Implicit Reference
        & $98.93 \pm 0.24$
        & $0.0659 \pm 0.0000$
        & $0.1303 \pm 0.0003$
        & $0.2407 \pm 0.0001$
        & $0.1470 \pm 0.0003$
        & $0.1239 \pm 0.0001$ \\

        SP-GMT w/o Robot Pelvis Pose
        & $99.33 \pm 0.27$
        & $0.0356 \pm 0.0000$
        & $\underline{0.0753} \pm 0.0001$
        & $0.1330 \pm 0.0001$
        & $\underline{0.0836} \pm 0.0002$
        & $0.0663 \pm 0.0001$ \\

        SP-GMT (Ours)
        & $99.83 \pm 0.08$
        & $\underline{0.0278} \pm 0.0000$
        & $\mathbf{0.0676} \pm 0.0002$
        & $\underline{0.0889} \pm 0.0001$
        & $\mathbf{0.0733} \pm 0.0001$
        & $0.0516 \pm 0.0000$ \\

        SONIC~\cite{luo2026sonic}
        & $\mathbf{100.00} \pm 0.00$
        & $0.0294 \pm 0.0000$
        & $0.4177 \pm 0.0015$
        & $0.1097 \pm 0.0000$
        & $0.4166 \pm 0.0015$
        & $\underline{0.0509} \pm 0.0000$ \\

        Humanoid-GPT~\cite{qi2026humanoidgpt}
        & $\underline{99.93} \pm 0.08$
        & $\mathbf{0.0245} \pm 0.0000$
        & $0.4742 \pm 0.0052$
        & $\mathbf{0.0572} \pm 0.0000$
        & $0.4778 \pm 0.0052$
        & $\mathbf{0.0434} \pm 0.0000$ \\

        \midrule

        \multicolumn{7}{l}{\textit{BONES-SEED Diverse-500}} \\

        SP-GMT w/ Joint-Space Reference

        & $90.21 \pm 0.10$
        & $0.0315 \pm 0.0001$
        & $\underline{0.0906} \pm 0.0008$
        & $0.0844 \pm 0.0002$
        & $\underline{0.0908} \pm 0.0008$
        & $0.0568 \pm 0.0002$ \\

        SP-GMT w/o Sparse Reference
        & $92.23 \pm 0.16$
        & $0.0418 \pm 0.0002$
        & $0.5763 \pm 0.0416$
        & $0.1380 \pm 0.0004$
        & $0.5804 \pm 0.0414$
        & $0.0772 \pm 0.0004$ \\

        SP-GMT w/o Implicit Reference
        & $91.19 \pm 0.20$
        & $0.0645 \pm 0.0001$
        & $0.1627 \pm 0.0009$
        & $0.2059 \pm 0.0002$
        & $0.1678 \pm 0.0009$
        & $0.1209 \pm 0.0002$ \\

        SP-GMT w/o Robot Pelvis Pose
        & $92.15 \pm 0.17$
        & $0.0401 \pm 0.0001$
        & $0.0958 \pm 0.0011$
        & $0.1309 \pm 0.0005$
        & $0.1025 \pm 0.0011$
        & $0.0737 \pm 0.0003$ \\

        SP-GMT (Ours)
        & $\underline{92.24} \pm 0.09$
        & $\underline{0.0299} \pm 0.0001$
        & $\mathbf{0.0797} \pm 0.0009$
        & $\underline{0.0810} \pm 0.0003$
        & $\mathbf{0.0798} \pm 0.0009$
        & $\underline{0.0553} \pm 0.0003$ \\

        SONIC~\cite{luo2026sonic}
        & $\mathbf{93.84} \pm 0.09$
        & $0.0386 \pm 0.0000$
        & $0.3181 \pm 0.0031$
        & $0.1091 \pm 0.0001$
        & $0.3188 \pm 0.0031$
        & $0.0609 \pm 0.0000$ \\

        Humanoid-GPT~\cite{qi2026humanoidgpt}
        & $90.10 \pm 0.12$
        & $\mathbf{0.0272} \pm 0.0001$
        & $0.5972 \pm 0.0108$
        & $\mathbf{0.0573} \pm 0.0001$
        & $0.5987 \pm 0.0108$
        & $\mathbf{0.0502} \pm 0.0001$ \\

        \bottomrule
    \end{tabular*}
\end{table*}

\begin{table*}[!t]
\centering
\begin{threeparttable}
\caption{Quantitative comparison of motion generation
quality and efficiency. P95 denotes the 95th percentile
of the corresponding error or jerk distribution.
Reporting and highlighting conventions follow
Table~\ref{tab:motion-tracking-trajectory-noise}.}
\label{tab:motion_generation_ablation}
\footnotesize
\setlength{\tabcolsep}{3pt}
\renewcommand{\arraystretch}{1.15}
\begin{tabular*}{\textwidth}{@{\extracolsep{\fill}}lcccccc@{}}
\toprule
\multirow{2}{*}{Model}
& Network Depth
& EE Error Mean
& EE Error P95
& EE Jerk P95
& Body Jerk P95
& Generation Rate \\
& $\downarrow$ (layers)
& $\downarrow$ (cm)
& $\downarrow$ (cm)
& $\downarrow$ ($\mathrm{m/s^3}$)
& $\downarrow$ ($\mathrm{m/s^3}$)
& $\uparrow$ (Hz) \\
\midrule
Kimodo (20 steps)
& 16
& $1.941 \pm 1.398$
& $3.521 \pm 2.832$
& $\mathbf{450.768} \pm 224.996$
& $\mathbf{370.225} \pm 178.458$
& $5.405$ \\[2pt]
SP-MGM (w/o Stage II)
& 8
& $8.441 \pm 1.680$
& $12.772 \pm 3.080$
& $1068.626 \pm 506.355$
& $916.724 \pm 476.011$
& $7.634$ \\
SP-MGM (w/o $L_{\mathrm{temp}}$)
& 8
& $\underline{1.915} \pm 0.730$
& $\underline{3.420} \pm 1.445$
& $657.125 \pm 311.392$
& $570.246 \pm 287.323$
& $7.322$ \\
SP-MGM (Ours)
& 8
& $\mathbf{1.812} \pm 0.709$
& $\mathbf{3.212} \pm 1.289$
& $\underline{514.337} \pm 236.943$
& $\underline{416.945} \pm 200.507$
& $\underline{7.813}/\mathbf{11.494}$ \\
\bottomrule
\end{tabular*}
\begin{tablenotes}[flushleft]
\footnotesize
\item[] Network depth denotes the number of transformer blocks in the motion generator.
For our SP-MGM model, 7.813~Hz and 11.494~Hz denote
the generation rates under standard inference and text-free inference
with TensorRT acceleration, respectively.
\end{tablenotes}
\end{threeparttable}
\end{table*}

As shown in Table~\ref{tab:motion-tracking-trajectory-noise}, among the compared methods, SP-GMT achieves the lowest global MPJPE and global EE error on both datasets. For local tracking, it ranks second in five of the six dataset--metric comparisons across local MPJPE, joint error, and local EE error. Notably, SP-GMT is trained on a substantially smaller motion corpus with only 192 GPU-hours, compared with 21k GPU-hours for SONIC and 15k GPU-hours for Humanoid-GPT. These results indicate competitive local motion fidelity and strong global trajectory tracking with substantially less data and training compute.

\subsubsection{Ablation Study of SP-GMT}
The ablations suggest that the explicit and implicit references provide complementary information. Using only the latent reference (w/o Sparse Reference) leads to substantially larger global tracking errors, while using only the five-point reference (w/o Implicit Reference) degrades local tracking and joint accuracy. Replacing the latent reference with a joint-space reference also increases global tracking error. Combining sparse spatial targets with the learned latent reference in SP-GMT yields the lowest global MPJPE and EE error on both benchmarks while remaining competitive in local tracking, demonstrating the benefit of the explicit--implicit representation.

Removing the robot pelvis pose causes only a moderate performance drop compared with the full model. This suggests that SP-GMT is not strongly dependent on precise pelvis localization.

\subsection{Motion Generation Quality and Efficiency}
\label{sec:motion_generation_results}

We select 600 motions from BONES-SEED~\cite{rempe2026kimodo}, covering diverse motion dynamics and spatial extent. Sparse-pose adherence is measured by EE error, and motion smoothness by EE jerk and body jerk.
Lower error and jerk values indicate better motion
generation quality, while a higher generation rate indicates
greater efficiency. All results are summarized in Table~\ref{tab:motion_generation_ablation}.

\subsubsection{Comparison with Motion-Generator Baseline}
Under the same 20-step denoising setting, we compare SP-MGM
against the uncompressed Kimodo baseline. Despite modest
increases in EE jerk and body jerk, SP-MGM achieves lower mean
and P95 EE errors. Notably, its generation rate reaches up to 1.45$\times$ that of Kimodo.  These results demonstrate that our two-stage distillation
strategy enables more accurate and efficient motion
generation under sparse-pose constraints.

\subsubsection{Ablation study of SP-MGM}
Full SP-MGM achieves the best performance across all metrics among the compared SP-MGM variants.
With Stage I alone (w/o Stage II), SP-MGM exhibits larger EE errors
and higher body jerk, with P95 EE error increasing from 3.212 to
12.772\,cm. These results indicate that Stage I establishes basic
motion generation capability, while Stage II substantially improves
adherence to the sparse-pose targets and temporal smoothness.
Without $L_{\mathrm{temp}}$, EE accuracy remains close
to that of full SP-MGM, but EE and body jerk P95 increase by 27.8\%
and 37.0\%, respectively. Temporal regularization thus suppresses
high-frequency jitter while also slightly reducing EE errors.

\subsection{Loco-Manipulation Evaluation}

\subsubsection{Experimental Setup}
We evaluate the complete SP-HLM system in both simulation and the real world. To provide a wider field of view, we equip the Unitree G1 with a ZED 2i stereo camera.

Our simulated evaluation consists of eight loco-manipulation tasks, as illustrated in Fig.~\ref{fig:simulated-task-keyframes}.
For the six tasks designed in-house (T1--T6), we collect 120 teleoperation demonstrations
per task using SONIC. Following the offline replay-and-render paradigm of
SIMPLE~\cite{wei2026simple}, we augment these demonstrations by randomizing
object positions, robot initial positions, and lighting conditions.
To avoid restricting the evaluation to our own
demonstrations, we include two additional loco-manipulation tasks from the HumanoidArena benchmark~\cite{wang2026humanoidarena} (T7 and T8), together with
their benchmark demonstrations.
The resulting eight tasks cover whole-body workspace extension, body-part
interaction, and locomotion-conditioned manipulation.

For the real-world evaluation, we select four representative
loco-manipulation tasks in everyday settings, as shown in
Fig.~\ref{fig:real-world-tasks}. We collect 80 teleoperation demonstrations per task using
SONIC, whereas SP-HLM employs SP-GMT for whole-body control
during evaluation. Thus, this
setting also evaluates whether the
sparse-pose-driven interface preserves task intent across
different whole-body trackers.

\subsubsection{Simulated Loco-Manipulation}

We evaluate each method over 50 trials per task. For each evaluation trial, the object position, robot initial position, and lighting condition are independently resampled. Both GR00T baselines use the same SONIC backend, making the comparison more reflective of differences in motion interface. As shown in Table~\ref{tab:sim_success}, SP-HLM achieves the highest overall success rate among the evaluated methods. This result highlights the advantage of the
sparse-pose-driven interface over the evaluated token-space and joint-space interfaces for coordinated whole-body
loco-manipulation. In the \textit{Direct Latent Prediction} variant, SP-VLA
directly predicts the motion latent alongside sparse poses and finger joint-position targets. The overall success rate drops to 23.25\%, suggesting that separating explicit spatial targets from implicit whole-body motion generation is beneficial for loco-manipulation. 
Fig.~\ref{fig:simulated-task-keyframes} shows SP-HLM execution keyframes and
language instructions for all eight simulated tasks.

\subsubsection{Real-World Loco-Manipulation} 
\begin{figure}[!htbp]
    \vspace*{2mm}
    \centering
    \begin{minipage}{0.48\linewidth}
        \centering
        \includegraphics[width=\linewidth]{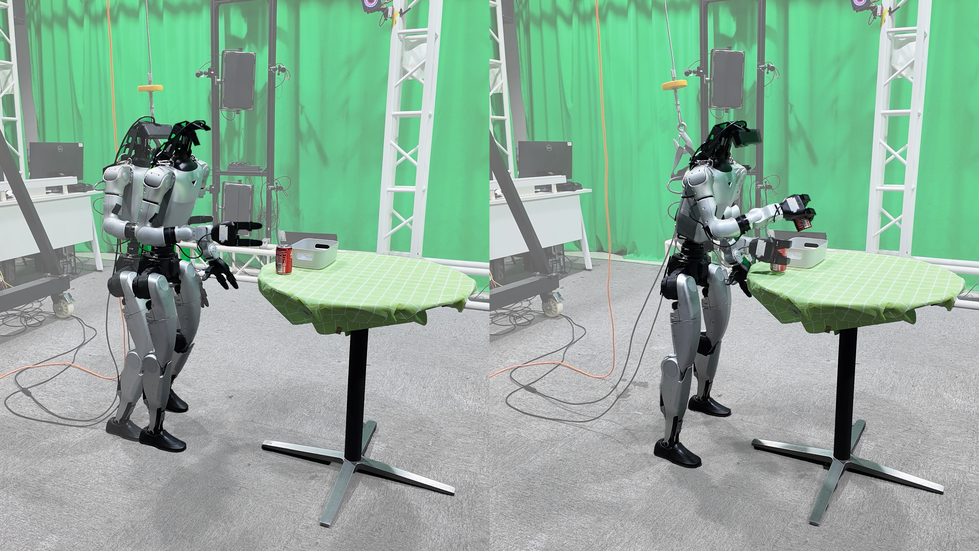}
        \par\small T1: Object Placement
    \end{minipage}
    \hfill
    \begin{minipage}{0.48\linewidth}
        \centering
        \includegraphics[width=\linewidth]{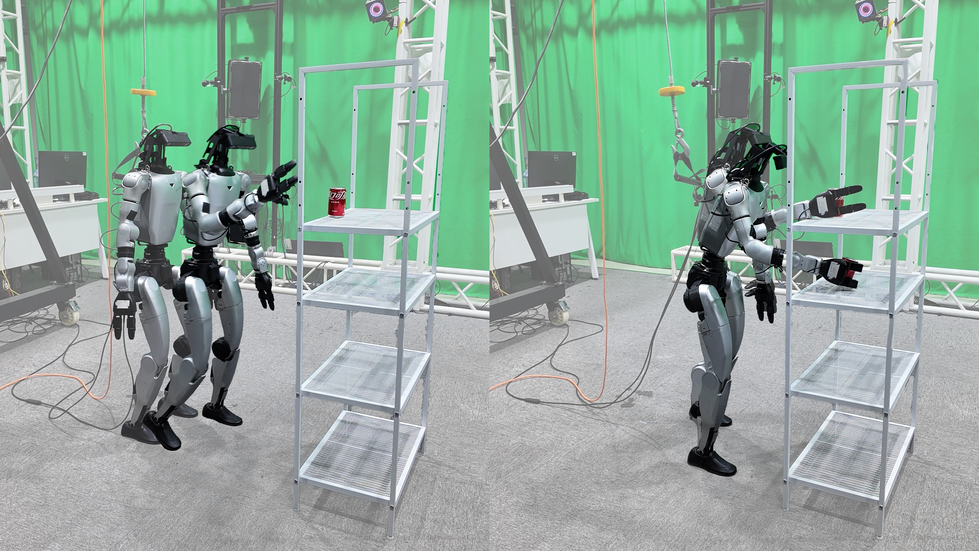}
        \par\small T2: Shelf Transfer
    \end{minipage}
    \par\vspace{4pt}
    \begin{minipage}{0.48\linewidth}
        \centering
        \includegraphics[width=\linewidth]{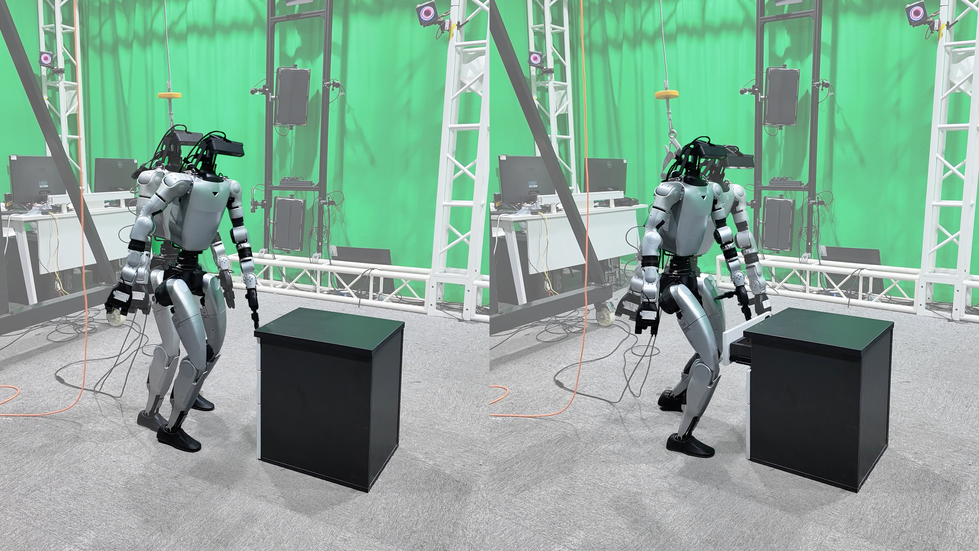}
        \par\small T3: Cabinet Opening
    \end{minipage}
    \hfill
    \begin{minipage}{0.48\linewidth}
        \centering
        \includegraphics[width=\linewidth]{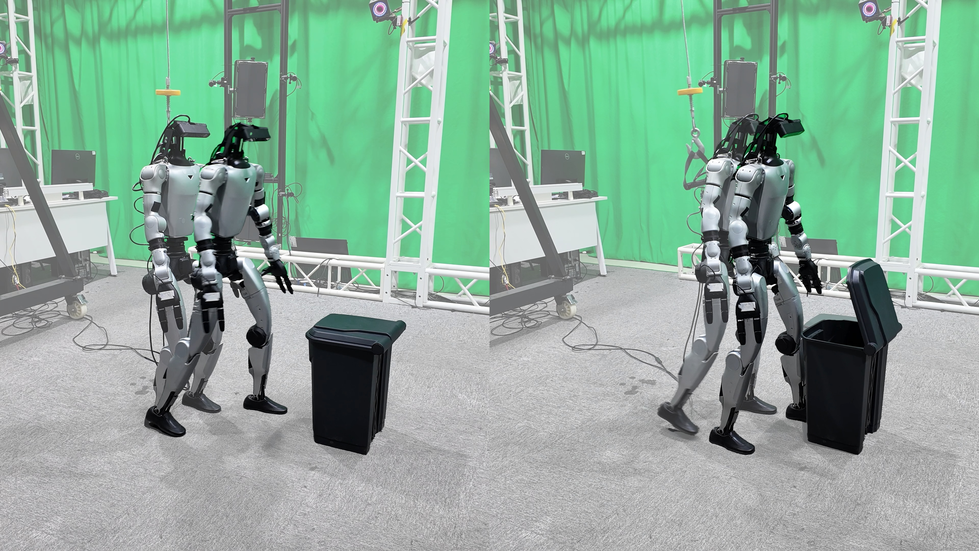}
        \par\small T4: Trash-Bin Opening
    \end{minipage}
    \caption{Representative rollouts of SP-HLM on four real-world humanoid loco-manipulation tasks.}
    \label{fig:real-world-tasks}
\end{figure}

To further assess deployment performance, we
evaluate SP-HLM over 20 trials per real-world task. As shown in Table~\ref{tab:real-world-deployment-success},
SP-HLM achieves success rates exceeding 50\% on both object placement
and shelf transfer. Even on the more challenging cabinet opening and trash-bin opening tasks, it achieves success rates of
25\% and 40\%, respectively. Fig.~\ref{fig:real-world-tasks} shows representative
execution keyframes from the real-world deployments.


\begin{table}[!htbp]
    \centering
    \caption{Real-world success rates of SP-HLM.}
    \label{tab:real-world-deployment-success}
    \footnotesize
    \setlength{\tabcolsep}{2pt}
    \renewcommand{\arraystretch}{1.15}
    \begin{tabularx}{\linewidth}{@{\hspace{4pt}}l*{4}{>{\centering\arraybackslash}X}@{\hspace{4pt}}}
        \toprule
        Real-world Task & T1: OP & T2: ST & T3: CO & T4: TO \\
        \midrule
        Success Rate & 13/20 & 11/20 & 5/20 & 8/20 \\
        \bottomrule
    \end{tabularx}
\end{table}

\paragraph*{Failure Case Analysis}
The largest performance drops relative to simulation occur
on tasks requiring precise manipulation under demanding
balance conditions.
Cabinet opening requires aligning the
hand with a small handle while maintaining a single-leg squat, whereas
trash-bin opening requires lifting one foot and placing it
precisely on the narrow pedal. In both tasks, even slight body sway
can induce end-effector pose
deviations, compromising contact
accuracy and leading to task failure.

\FloatBarrier

%% file: content/conclusion.tex
\section{CONCLUSIONS}
We presented SP-HLM, a humanoid
loco-manipulation framework built on a sparse-pose-driven interface
that integrates explicit sparse-pose constraints with an
implicit full-body motion prior. Three components realize this interface: SP-VLA predicts five-point poses and hand commands,
SP-MGM reconstructs coordinated full-body motion, and SP-GMT integrates the explicit and implicit
commands for whole-body control. Experiments on motion tracking, motion generation, and loco-manipulation in both simulation and the real world demonstrate the effectiveness
of the proposed interface and the complete framework.

However, this study focuses primarily on whole-body coordination. Hand control is specified solely through finger joint-position targets, and we do not explore alternative hand-level interfaces or dexterous manipulation. Extending the proposed interface to support whole-body dexterous manipulation is an important direction for future work.

%% file: content/appendix.tex
\subsection{SP-GMT Training Details}
We adopt the reward formulation of
BeyondMimic~\cite{liao2025beyondmimic}, augmented with
global position and orientation tracking rewards for
four end effectors (both wrists and ankles), as well as
penalties for undesired foot motions.
The reward terms and their default weights are
summarized in Table~\ref{tab:tracking_rewards}.
\begin{table}[htbp]
\centering
\caption{Reward terms and default weights.}
\label{tab:tracking_rewards}
\begingroup
\footnotesize
\setlength{\tabcolsep}{3pt}
\renewcommand{\arraystretch}{1.08}
\begin{tabularx}{\columnwidth}{@{}Xrr@{}}
\toprule
Reward term & Weight & $\sigma$ \\
\midrule
Global pelvis position          & $0.5$   & $0.3$ m \\
Global pelvis orientation       & $0.5$   & $0.4$ rad \\
Relative body position          & $1.0$   & $0.3$ m \\
Relative body orientation       & $1.0$   & $0.4$ rad \\
Global body linear velocity     & $1.0$   & $1.0$ m/s \\
Global body angular velocity    & $1.0$   & $3.14$ rad/s \\
Global end-effector position    & $1.0$   & $0.3$ m \\
Global end-effector orientation & $1.0$   & $0.4$ rad \\
\midrule
Action smoothness, $\|\Delta a_t\|_2^2$
                                & $-0.1$  & -- \\
Joint-limit violation           & $-10.0$ & -- \\
Undesired contacts              & $-0.1$  & -- \\
Foot sliding                    & $-0.3$  & -- \\
Self-collision                  & $-1.0$  & -- \\
Height-weighted foot air time   & $4.0$   & -- \\
\bottomrule
\end{tabularx}

\smallskip
\parbox{\columnwidth}{\scriptsize
$\Delta a_t=a_t-a_{t-1}$.
Undesired contacts exclude the wrists and ankles.
Contact and self-collision thresholds are 1\,N.
Foot sliding sums horizontal foot speeds during contact
and is gated by $p_{\mathrm{fail}}<1$.
The air-time term is non-positive:
$\sum_f \min(T_f-0.8,0)$ at touchdown, where $T_f$
accumulates time with a height-dependent multiplier
in $[0.2,1.0]$ over foot heights $[0.05,0.15]$\,m.
It is disabled for randomly sampled static segments.
Zero-weight and disabled terms are omitted.}
\endgroup
\end{table}
\subsection{SP-MGM Training Details}
\label{app:spmgm-training}

SP-MGM is initialized from the pretrained KIMODO motion generator and
compresses both the network depth and the number of denoising steps.
Stage I performs offline distillation on paired motion data, jointly
supervising the student with predictions from the frozen teacher and
ground-truth motion. Stage II performs rollout-aware distillation on
student-visited states: intermediate rollout states are relabeled by the
frozen teacher, and temporal regularization is applied to improve motion
smoothness. The detailed training configurations are listed in
Table~\ref{tab:spmgm_training}.

\begin{center}
    \centering
    \captionof{table}{Hyperparameters for SP-MGM training.}
    \label{tab:spmgm_training}
    \footnotesize
    \setlength{\tabcolsep}{3pt}
    \renewcommand{\arraystretch}{1.12}
    \begin{tabularx}{\columnwidth}{@{}X>{\raggedleft\arraybackslash}p{0.40\columnwidth}@{}}
        \toprule
        Hyperparameter & Configuration \\
        \midrule
        \multicolumn{2}{@{}l}{\textit{Student Model}} \\
        Root / Body Transformer Layers & 8 / 8 \\
        Hidden / Feed-Forward Dimensions & 1024 / 2048 \\
        Number of Attention Heads & 8 \\
        Diffusion / DDIM Sampling Steps & 1000 / 20 \\
        \midrule
        \multicolumn{2}{@{}l}{\textit{Shared Optimization}} \\
        Optimizer & Adam-atan2 \\
        Optimizer Betas $(\beta_1,\beta_2)$ & $(0.9, 0.999)$ \\
        \midrule
        \multicolumn{2}{@{}l}{\textit{Stage I: Offline Distillation}} \\
        Training Steps / Global Batch Size & 10{,}000 / 128 \\
        Learning Rate & $2\times10^{-5}$ \\
        EMA Decay & 0.995 \\
        Teacher / GT Weights (Steps 0--8{,}000) & $0.9/0.1\rightarrow0.3/0.7$, linear \\
        Teacher / GT Weights (Steps 8{,}000--10{,}000) & $0.3/0.7$, fixed \\
        \midrule
        \multicolumn{2}{@{}l}{\textit{Stage II: On-Policy Distillation}} \\
        Training Steps / Global Rollout Batch Size & 100{,}000 / 16 \\
        Sampled States per Rollout & 3 \\
        Teacher / GT Weights & $0.7/0.3$ \\
        Base / Minimum Learning Rate & $2\times10^{-5}/2\times10^{-6}$ \\
        Learning Rate Schedule & Cosine decay \\
        EMA Decay & 0.997 \\
        Self / Root / Heading Smoothness Weights & $5.0/1.0/1.0$ \\
        \bottomrule
    \end{tabularx}
\end{center}

\subsection{SP-VLA Training and Inference Details}
\label{app:sp-vla-training}

The SP-VLA policy is fine-tuned from GR00T N1.7 for 160{,}000
optimization steps with a global batch size of 32 and a learning rate of
$1\times10^{-4}$. It predicts 40-step action chunks using a 59-D representation comprising 14 hand-joint values, a 9-D root pose,
and four 9-D end-effector poses. Prefix-conditioned RTC is enabled during
training, with delays randomly sampled from 0 to 8 steps. During inference,
the first 34 actions are executed and the final 6 actions are retained as
the clean prefix for the next prediction, forming a rolling-horizon scheme. Action generation uses four flow-matching
inference steps.